\documentclass[lettersize,journal]{IEEEtran}
\usepackage{amsmath,amsfonts}
\usepackage{algorithmic}
\usepackage{algorithm}
\usepackage{array}
\usepackage[caption=false,font=normalsize,labelfont=sf,textfont=sf]{subfig}
\usepackage{textcomp}
\usepackage{stfloats}
\usepackage{url}
\usepackage{verbatim}
\usepackage{graphicx}
\usepackage{multirow}
\usepackage{cite}
\usepackage[colorlinks,linkcolor=blue,anchorcolor=blue, citecolor=blue]{hyperref}
\usepackage{xcolor}

\providecommand{\rev}[1]{\textcolor{red}{#1}}
\providecommand{\new}[1]{\textcolor{blue}{#1}}
\renewcommand{\rev}[1]{#1}
\renewcommand{\new}[1]{#1}

\usepackage{chngcntr}
\counterwithout{figure}{section}

\graphicspath{{Figure/}}
\usepackage{booktabs}
\usepackage{amsmath}
\usepackage{microtype}
\usepackage{algorithm} 
\usepackage{algorithmic}

\begin{document}

	\title{Learning A Physical-aware Diffusion Model Based on Transformer for Underwater Image Enhancement}
	
	\author{Chen Zhao$^{*}$, Chenyu Dong$^{*}$, Weiling Cai~\IEEEmembership{Member,~IEEE}, Yueyue Wang
		
		\thanks{This work was supported by the National Natural Science Foundation of China (Grant No.
62276138 and 62371232). $^{*}$ indicates that Chen Zhao and Chenyu Dong contributed equally to this work. \textit{(Corresponding author: Weiling Cai).}
		}
		\thanks{
			Chen Zhao are with the School of Intelligence Science and Technology, Nanjing University,  Suzhou, Jiangsu, China (E-mail: 602024710020@smail.nju.edu.cn).     
            }
            \thanks{
            Chenyu Dong, Yueyue Wang and Weiling Cai are with the School of Artificial Intelligence, Nanjing Normal University, Nanjing, Jiangsu, China (E-mail: 546845583@qq.com, 3385839548@qq.com, caiwl@njnu.edu.cn).
            }

	}
	
	\maketitle
	
	\begin{abstract}
		Underwater visuals undergo various complex degradations, inevitably influencing the efficiency of underwater vision tasks. Recently, diffusion models were employed to underwater image enhancement (UIE) tasks, and gained the best performance. However, these methods fail to consider the physical properties and underwater imaging mechanisms in the diffusion process, limiting information completion capacity of diffusion models. \rev{ In this paper, we introduce a novel UIE framework, named PA-Diff, designed to exploit physical knowledge to guide the diffusion process.}
	\rev{PA-Diff consists of the Physics Prior Generation (PPG) branch, the Implicit Neural Reconstruction (INR) branch, and the Physics-aware Diffusion Transformer (PDT) branch.} Our designed PPG branch aims to produce the prior knowledge of physics. With utilizing the physics prior knowledge to guide the diffusion process, PDT branch can obtain underwater-aware ability and model the complex distribution in real-world underwater scenes. INR Branch can learn robust feature representations from diverse underwater images via implicit neural representation, which reduces the difficulty of restoration for PDT branch. \rev{Extensive experiments demonstrate that our method achieves the best performance on UIE tasks.}  The code is available at \href{https://github.com/chenydong/PA-Diff}{https://github.com/chenydong/PA-Diff}.

	\end{abstract}
	
	\begin{IEEEkeywords}
		Underwater image enhancement, Diffusion model, Physical model.
	\end{IEEEkeywords}
	
	\section{Introduction}
	\IEEEPARstart{T}{he} primary objective of underwater image enhancement (UIE) lies in the attainment of superior-quality images, achieved through the elimination of scattering effects and rectification of color distortions prevalent in degraded images. UIE stands as an indispensable component for tasks related to vision in the underwater domain.
In recent years, UIE has been intensively studied, and many methods are proposed \cite{mu2023generalized}\cite{TangKI23}\cite{Zhaowfdiff}. \rev{Mathematically, most of these methods are characterized by the simplified atmospheric model}:
\begin{equation}
    I^c(x)=J^c(x)T^c(x)+(1-T^c(x))B^c,
\end{equation}
where $c\in\{R, G, B\}$ is the color channel, $I^c(x)$ is the underwater image, $J^c(x)$ is the scene direct reflected light at $x$, $t^c(x)$ is \rev{the medium transmission map}, $B^c$ is  the global background scattered light. The goal of the UIE is to recover the clear $J^c(x)$ from   \rev{a given underwater image $I^c(x)$}.

Conventional UIE approaches, which hinge on the intrinsic characteristics of underwater images, have been presented \cite{PengC17,DrewsNMBC13,PengCC18}. These techniques delve into the physical aspects contributing to degradation, such as color cast or scattering, seeking to rectify and enhance the underwater images. Nevertheless, these models grounded in physics exhibit restricted representational capacities, particularly in the face of highly intricate and diverse underwater scenarios. 

To address this problem, many learning-based methodologies \cite{TangKI23,PengZB23,LiGRCHKT20,FabbriIS18} have been introduced to yield superior results. Leveraging the potent feature representation and nonlinear mapping capabilities of neural networks, these methods excel. Those  methods 
can be roughly categorized into two classes: physics-free end-to-end methods \cite{jiang2022two,PengZB23}
and physics-aware methods \cite{LiAHCGR21,mu2023generalized,fu2022unsupervised}. 
\rev{ Regarding physics-free end-to-end methods, although they achieve decent performance, they mainly operate in the raw image space, which does not fully explore the useful physics-aware feature information for UIE. Moreover, the interpretability of physics-free models remains underexplored regarding the physical process of UIE.}
Regarding the physics-aware deep methods, most of them utilize the atmospheric scattering model. Existing UIE approaches \cite{LiAHCGR21,fu2022unsupervised} develop deep neural networks to estimate the transmission and the background scattered light, and then follow the traditional UIE methods to restore underwater images. However, these approaches do not correct the errors of the atmospheric light estimated by neural networks, which seriously affects the quality of the images \cite{ye2022underwater}. Therefore, there is a crucial problem to adequately exploit the physical information and effectively incorporate them into a unified UIE deep framework.

\begin{figure*}[t]
	\centering
	\includegraphics[width=1\linewidth]{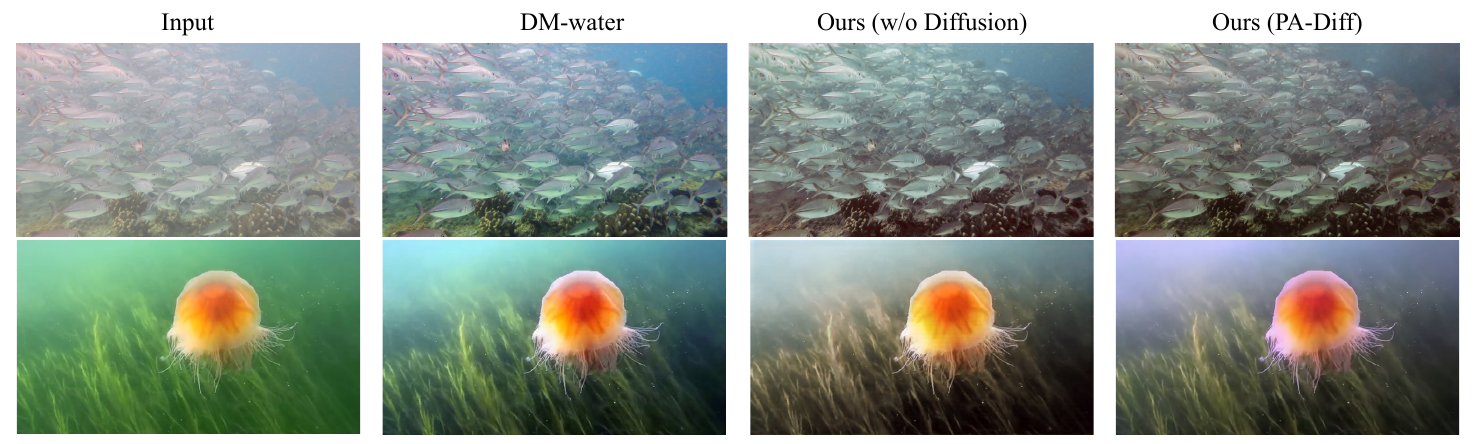}

	\caption{Visualization comparison of 1080P images in real-world scenarios. Please zoom to view. Compared to DM-Water \cite{TangKI23}, both  \rev{ our full model and its version without diffusion technique} exhibit stronger generalization capability for real-world underwater image enhancement, demonstrating the crucial role of physical priors in improving model generalization. Moreover, with the diffusion technique, our PA-Diff produces clearer results, further validating the effectiveness of the diffusion model.}
 \label{fig:motivation}

\end{figure*}

	Recently, there has been a surge of interest in image synthesis \cite{lu2024mace,lu2023tf,zhou2024migc} and restoration tasks \cite{ChoiKJGY21,WangYZ23, abs-2308-13164,zhou2023pyramid}, with a notable focus on diffusion-based techniques like DDPM \cite{DDPM} and DDIM \cite{DDIM}. This heightened attention can be attributed to the remarkable generative capabilities of diffusion models. Tang et al. \cite{TangKI23} present an underwater image enhancement approach with diffusion model, WF-Diff \cite{Zhaowfdiff} proposed \rev{a frequency diffusion model} for UIE, and UIEDP proposed a UIE framework with Diffusion Prior\cite{du2023uiedp}. 
	However, these methods fail to consider the physical properties and underwater imaging mechanisms in the diffusion process, limiting their information completion capacity for the diffusion model. Moreover, the diffusion models lack awareness regarding the difficulty of enhancing different image regions, a critical consideration for modeling the complex distribution in real-world underwater scenes. As shown in Figure \ref{fig:motivation}, compared to previous diffusion-based method \cite{TangKI23}, our approach achieves superior generalization capability in real-world scenarios.

	In this paper, we develop a novel physics-aware diffusion model, fully exploiting physical information to guide the diffusion process, called PA-Diff. It mainly consists of three branches: Physics Prior Generation (PPG) Branch, Implicit Neural Reconstruction (INR) Branch, and Physics-aware Diffusion Transformer (PDT) Branch. The PPG branch aims to generate the transmission map and the global background light as physics prior information. The transmission map is exploited as the confidence guidance for the PDT branch which enables our PA-Diff with underwater-aware ability. The global background light can be regarded as a prior condition on both the type and degree of degradation for the PDT Branch. Inspired by the latest works \cite{NeRD-Rain, yang2023implicit} based on INR that can encode an image as a continuous function and learn a more robust degradation representation, 
	we further incorporate the INR into our proposed framework to \rev{learn robust feature representations from diverse underwater images.} Finally, the PDT branch employs the strong generation ability of diffusion models to restore underwater images, with the guidance of physical information. Specifically, \rev{we design a physics-aware diffusion transformer block which contains a physics-aware self-attention  (PA-SA)}, physics perception unit (PPU) and gated multi-scale feed-forward network  (GM-FFN) to exploit physics prior information and capture the long-range diffusion dependencies.  Our PA-Diff achieves unprecedented perceptual performance in UIE tasks. Extensive experiments demonstrate that our developed PA-Diff performs the superiority against previous UIE approaches, and \rev{ comprehensive ablation studies validate the effectiveness of each proposed component.}
	
	In summary, the main contributions of our PA-Diff are as follows:

	\begin{itemize}
		\item  \rev{ We propose a novel UIE framework based on physics-aware diffusion model, named PA-Diff, which consists of Physics Prior Generation (PPG) branch, Implicit Neural Reconstruction (INR) branch, and Physics-aware Diffusion Transformer (PDT) branch.  To the best of our knowledge, it is the first  diffusion model with the guidance of physical knowledge in UIE tasks}.
		\item \rev{We incorporate Implicit Neural Reconstruction (INR) into PA-Diff to learn robust feature representations from diverse underwater images, thereby reducing the difficulty of restoration for diffusion models}.
		\item We design a physics-aware diffusion transformer block, which not only enables PA-Diff with underwater-aware ability to guide the diffusion process, but also captures the long-range diffusion dependencies. \rev{Moreover, we propose a Physics Perception Unit (PPU) to leverage beneficial feature-level physical priors for underwater image enhancement.}
		\item  \rev{Extensive experiments against best methods demonstrate that our PA-Diff consistently outperforms previous UIE approaches}, and extensive ablation experiments can demonstrate the effectiveness of all contributions.
		
	\end{itemize}
	
	The remainder of this paper is organized as follows. Section II presents the related work, followed by a detailed description of our PA-Diff in Section III. Experimental results are provided in Section IV, and conclusions are drawn in Section V.

\begin{figure*}[t]
	\centering
	\includegraphics[width=1\linewidth]{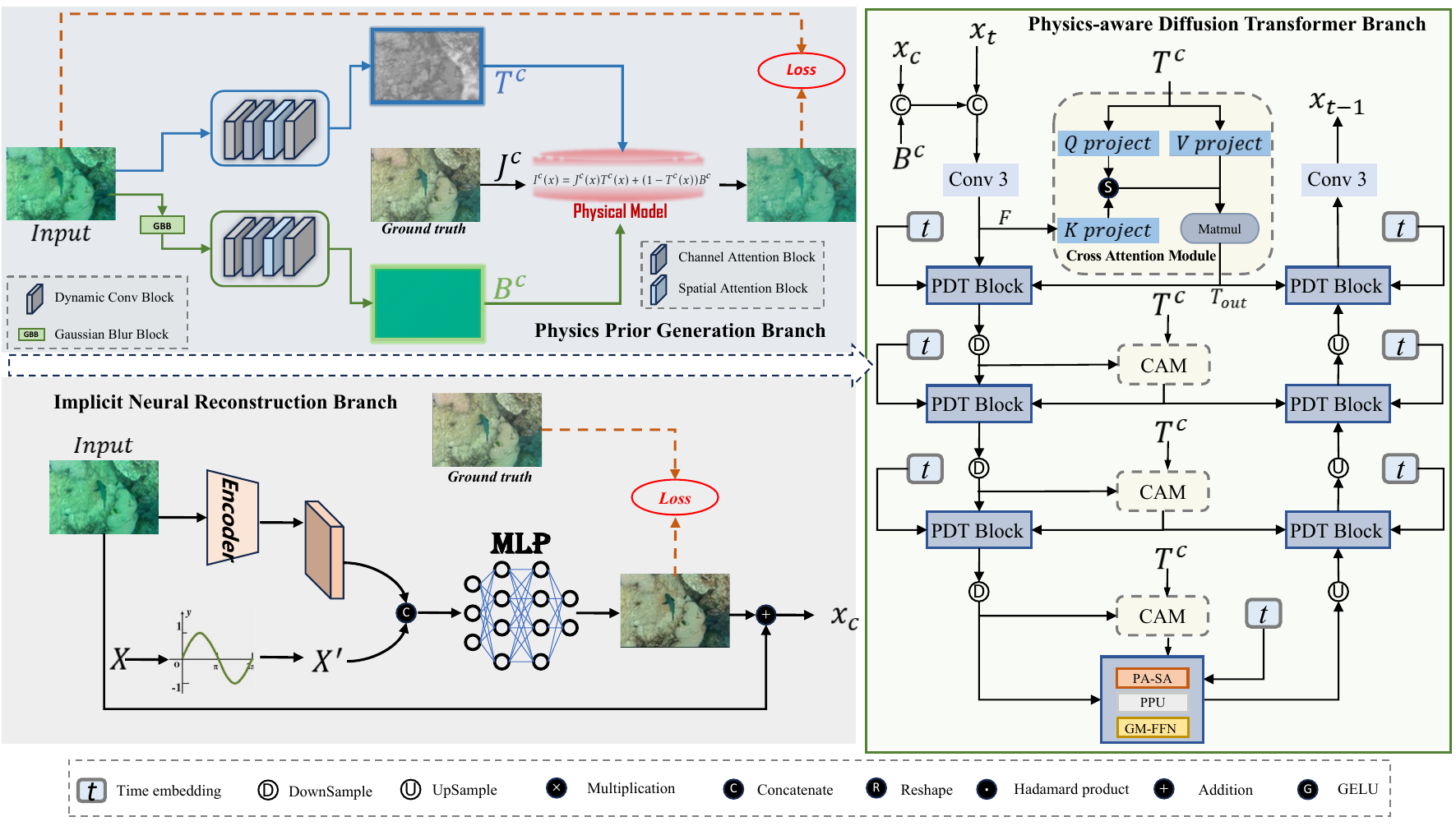}
	
	\caption{\rev{The overall framework of the proposed PA-Diff. A-Diff consists of three cooperative branches: the Physics Prior Generation (PPG) branch, the Implicit Neural Reconstruction (INR) branch, and the Physics-aware Diffusion Transformer (PDT) branch.}  The PPG branch aims to produce the prior knowledge of physics. With utilizing this prior to guide the diffusion process, the PDT branch can obtain underwater-aware ability and model the complex distribution in real-world underwater scenes. \rev{ INR Branch can learn robust feature representations from diverse underwater images, which reduces the difficulty of restoration for the diffusion models.}}
 \label{fig:frame}
	
\end{figure*}

    \section{Related Work}
	\subsection{Underwater Image Enhancement }
Currently, \rev{existing UIE methods} can be briefly categorized into
the physical and deep model-based approaches \cite{PengC17,PengCC18,TangKI23,PengZB23,LiGRCHKT20}. \rev{Most UIE methods } based on the physical model utilize prior knowledge to establish models, such as water dark channel priors \cite{PengC17}, attenuation curve priors \cite{WangLC18}, fuzzy priors \cite{ChiangC12}. However, the manually established priors restrain the model’s robustness and scalability under the complicated and varied circumstances. In recent years, underwater physical imaging models are gradually utilized in combination with data-driven methods \cite{LiAHCGR21}. Recently, deep learning-based methods \cite{TangKI23,PengZB23,LiGRCHKT20,zhao2024cycle,fu2022uncertainty} achieve acceptable performance, and some complex frameworks are proposed and \rev{ achieve state-of-the-art (SOTA) performance} \cite{PengC17,DrewsNMBC13,zhao2023toward}. Wang et al. \cite{wang2017deep} introduced UIE-Net, incorporating a pixel-level perturbation technique to mitigate subtle texture variations, thereby enhancing both convergence efficiency and learning precision. Li et al. \cite{LiGRCHKT20} developed Water-Net, utilizing algorithms like white balance, histogram equalization, and gamma correction to create diverse inputs for underwater imagery, refining them via a feature transformation module and merging them with confidence maps from a gated fusion framework to achieve the final output. Qi et al. \cite{qi2023deep} separated the underwater image improvement process into color adjustment and visibility enhancement stages, formulating CCMSR-Net to extract both local attributes and global structures. Peng et al. \cite{PengZB23} presented a U-shaped Transformer combined with a multiscale feature fusion component and a spatially aware global feature unit, explicitly enhancing attention toward color channels and spatial zones undergoing severe degradation. 
\new{
WaterFormer \cite{wen2024waterformer} introduces a global–local Transformer to alleviate detail loss and incorporates a learnable Environment Adaptor to adaptively handle diverse underwater conditions.
SSD-UIE \cite{wen2025semi} presents a semi-supervised domain-adaptive framework to mitigate both inter-domain and intra-domain discrepancies. 
}
The previous methods \cite{PengZB23,LiGRCHKT20,DrewsNMBC13,zhao2023toward} neglect the underwater imaging mechanism and rely only on the
representation ability of deep networks. Ucolor \cite{LiAHCGR21} combined the underwater physical imaging model in the raw space and designed a medium transmission guided model. ATDCnet \cite{mu2023transmission} attempted to use transmission maps to guide deep neural networks. 
\new{
SyreaNet \cite{wen2023syreanet} adopts a physically guided generative adversarial framework to address the generalization challenge in real underwater image enhancement.
UIR-ES \cite{zhu2024uir} proposes a physics-inspired unsupervised framework to improve data-efficient learning and enhance generalization in real-world scenarios.
Therefore, how to fully exploit the knowledge of physics in deep neural networks is a very crucial problem. We aim to explore physics-aware diffusion framework, where physical priors are explicitly embedded into the diffusion process.}

\vspace{-5pt}
\subsection{Diffusion Model}
The fusion of deep learning with nonequilibrium thermodynamics has driven diffusion model advancements. 
Through denoising, the network extracts key data features by recovering original information from noise. Thus, the diffusion model not only reconstructs data structure but also forms a trainable generation framework \cite{DBLP:journals/tgrs/ZhangYC25}. Diffusion Probabilistic Models (DPMs) \cite{DDPM,DDIM} have been
widely adopted for conditional image generation  \cite{WangYZ23,lu2023tf,zhou2023pyramid,zhou2024migc,zhao2025ultrahr}. Currently, \rev{diffusion models have been applied to a variety of low-level visual tasks such as  image restoration \cite{ozdenizci2023restoring}, document enhancement\cite{yang2023docdiff}, and low light image enhancement \cite{yang2023difflle, zhou2023pyramid}.} The diffusion model provides benefits for handling intricate noise conditions in underwater visuals, owing to its strong noise removal ability and clear interpretability \cite{DBLP:journals/tgrs/ZhangYC25}.  Tang et al. \cite{TangKI23} present an underwater image enhancement approach with diffusion model and \rev{WF-Diff \cite{Zhaowfdiff} proposed a frequency diffusion model for UIE.} 
Lu et al. \cite{DBLP:journals/jvcir/LuGZL23} later presented an approach for underwater image improvement utilizing denoising diffusion probabilistic models (UW-DDPM). They also developed a real-time UIE strategy founded on the denoising diffusion probabilistic framework: SU-DDPM \cite{DBLP:journals/tcsv/LuGZL24}. Presently, some scholars are exploring diffusion models to refine existing enhancement techniques. Du et al. \cite{du2023uiedp} introduced UIE with diffusion prior (UIEDP), an innovative design that frames UIE as a posterior distribution sampling process, aiming to restore clear images from degraded underwater inputs. This method employs synthetic diffusion models to steer modern UIE algorithms, leading to enhanced results.  However, these methods fail to consider the physical properties and underwater imaging mechanisms in the diffusion process, limiting information completion capacity of diffusion models.

\section{Methodology}
	
\subsection{Overall Framework}
\rev{The overall framework of PA-Diff is shown in Figure \ref{fig:frame}. PA-Diff is designed to leverage the physical properties of underwater imaging mechanisms to guide the diffusion process, which 
which mainly consists of the Physics Prior Generation (PPG) branch, the Implicit Neural Reconstruction (INR) branch, and the Physics-aware Diffusion Transformer (PDT) branch.} The PPG branch aims to generate the transmission map and the global background light as physics prior information by utilizing the modified koschmieder light scanning model. The transmission map is exploited as the confidence guidance for the PDT branch which enables our PA-Diff with underwater-aware ability. The global background light can be regarded as a prior condition on both the type and degree of degradation for the PDT Branch. Inspired by \cite{ yang2023implicit}, based on implicit neural representations that can encode an image as a continuous function and learn a more robust feature representation, 
\rev{ we further introduce INR to learn robust degradation representations from diverse underwater images}, aiming to reduce the difficulty of restoration for diffusion models. The PDT branch employs the strong generation ability of diffusion models to restore underwater images, with the guidance of physical information. We adopt the diffusion process proposed in DDPM  \cite{DDPM} to construct the PDT branch, which can be described as a forward diffusion process and a reverse diffusion process.

\noindent\textbf{Forward Diffusion Process.} The forward diffusion process can be viewed as a Markov chain progressively adding
Gaussian noise to the data. Given a clean data $x_0$, then introduce Gaussian
noise based on the time step, as follows:  

\begin{equation}q(x_t|x_{t-1})=\mathcal{N}(x_t;\sqrt{1-\beta_t}x_{t-1},\beta_tI),\end{equation}
where $\beta_{t}$ is a variable controlling the variance of the noise. Introducing $\alpha_{t}=1-\beta_{t}$, this process can be described as:
\begin{equation}x_t=\sqrt{\alpha_t}x_{t-1}+\sqrt{1-\alpha_t}\epsilon_{t-1},\quad\epsilon_{t-1}\sim\mathcal{N}(0,\mathcal{Z}).\end{equation}

\rev{With Gaussian distributions are merged, we can obtain :}
\begin{equation}q(x_t|x_0)=\mathcal{N}(x_t;\sqrt{\bar{\alpha_t}}x_0,(1-\bar{\alpha_t})I).\end{equation}

\noindent\textbf{Reverse Diffusion Process.} The reverse diffusion process aims to restore the clean data from the Gaussian noise. The reverse diffusion can be expressed as:
\begin{equation}p_\theta(x_{t-1}|x_t,x_c)=\mathcal{N}(x_{t-1};\mu_\theta(x_t,x_c,x_p,t),\sigma_t^2\mathcal{Z}),\end{equation}
where $x_{c}$ refers to the conditional image ${I}$, and $x_{p}$ refers to the  physics prior. $\boldsymbol{\mu}_{\theta}(x_{t},x_{c},x_p,t)$ and $\sigma_t^2$ are the mean and variance from the estimate of
step t, respectively. We follow the setup of \cite{DDPM}, they can be expressed as:
\begin{equation}\mu_\theta(x_t,x_c,t)=\frac1{\sqrt{\alpha}_t}(x_t-\frac{\beta_t}{(1-\overline{\alpha}_t)}\epsilon_\theta(x_t,x_c,t)),\end{equation}

\begin{equation}\sigma_t^2=\frac{1-\overline{\alpha}_{t-1}}{1-\overline{\alpha}_t}\beta_t,\end{equation}
where $\epsilon_\theta(x_t,x_c,t)$ is the estimated value with a Unet.

We optimize an objective function for the noise estimated by the network and the noise $\epsilon$ actually added. Therefore,
The diffusion loss is:
\begin{equation}L_{dm}(\theta)=\|\epsilon-\epsilon_\theta(\sqrt{\overline{\alpha}_t}x_0+\sqrt{1-\overline{\alpha}_t}\epsilon,x_c,x_{p},t)\|.\end{equation}

\subsection{Physics Prior Generation}
\rev{Mathematically, the underwater images can be expressed by the simplified Koschmieder light scanning model \cite{akkaynak2019sea}:}
\begin{equation}
	I^c(x)=J^c(x)T^c(x)+(1-T^c(x))B^c,
\end{equation}
\rev{where $c\in\{R, G, B\}$ is the color channel, $I^c(x)$ is the observed underwater image intensity at pixel $x$ in channel $c$, consisting of both attenuated scene radiance and veiling-light backscatter; $J^c(x)$ is the latent clean radiance (clear-water equivalent), representing the true scene color if the medium were air instead of water; $T^c(x)$ is the transmission map, indicating the fraction of light that reaches the camera without being absorbed or scattered, depending on the scene depth and the wavelength-dependent attenuation coefficient; $B^c$ is the background light, modeling ambient light scattered into the camera from distant water volumes where object visibility diminishes. In the PPG branch, our goal is to produce $T^c(x)$ and $B^c$ from the $I^c(x)$. }

Network architecture of PPG branch is shown in Figure~\ref{fig:frame}, and consists of two sub-networks: transmission map generation sub-network $\mathcal{T}(x)$ and background light generation sub-network  $\mathcal{B}(x)$. The two sub-networks mainly consist of duplicated Dynamic Convolutions (DC) \cite{chen2020dynamic}.  DC dynamically combines numerous parallel convolutional kernels according to their attention, thereby enhancing the model's complexity and performance without augmenting the network's depth or breadth. Moreover, we introduce channel and spatial attention block to enhance the ability of feature learning.
 
Given that the background light differs from the transmission map and is independent of image content, it instead encapsulates global properties \cite{fu2022unsupervised}.
For the purpose of generating the background light component, 
our background light generation sub-network initially employs a Gaussian blur block (GBB) to eliminate fine-grained content details and thereby extract the inherent global attributes. Note that, for the background light generation sub-network, we first perform gaussian blur module to remove content details and obtain the global attribute. Different from the transmission map, the global background light is independent of the image content and reflects the global property. 

\rev{Given an underwater image as input $I$ and its corresponding ground truth (GT), we can obtain via Eq. 7:}
\begin{equation}
	\hat{I}=GT \cdot \mathcal{T}(I)+(1-\mathcal{T}(I))\mathcal{B}(GBB(I)),
\end{equation}

We constrain the PPG branch by supervised manner. The physics reconstruction loss of PPG branch can be expressed as:

\begin{equation}
	\mathcal{L}_p=||\hat{I}-I||_1.
\end{equation}

To enhance the perceptual similarity of the reconstructed images, we introduce the perceptual Loss $\mathcal{L}_{per}$ :
\begin{equation}
	\mathcal{L}_{per}=||\phi(\hat{I})-\phi(I)||_{2},
\end{equation}
where $\phi$ is a pre-trained VGG network.

\rev{ Finally, the overall loss $\mathcal{L}_{ppg}$ of the PPG branch can be expressed as:}
\begin{equation}
	\mathcal{L}_{ppg}=\lambda_{1}\mathcal{L}_{p} + \lambda_{2}\mathcal{L}_{per}. 
\end{equation}

\begin{figure*}[t]
	\includegraphics[width=\linewidth]{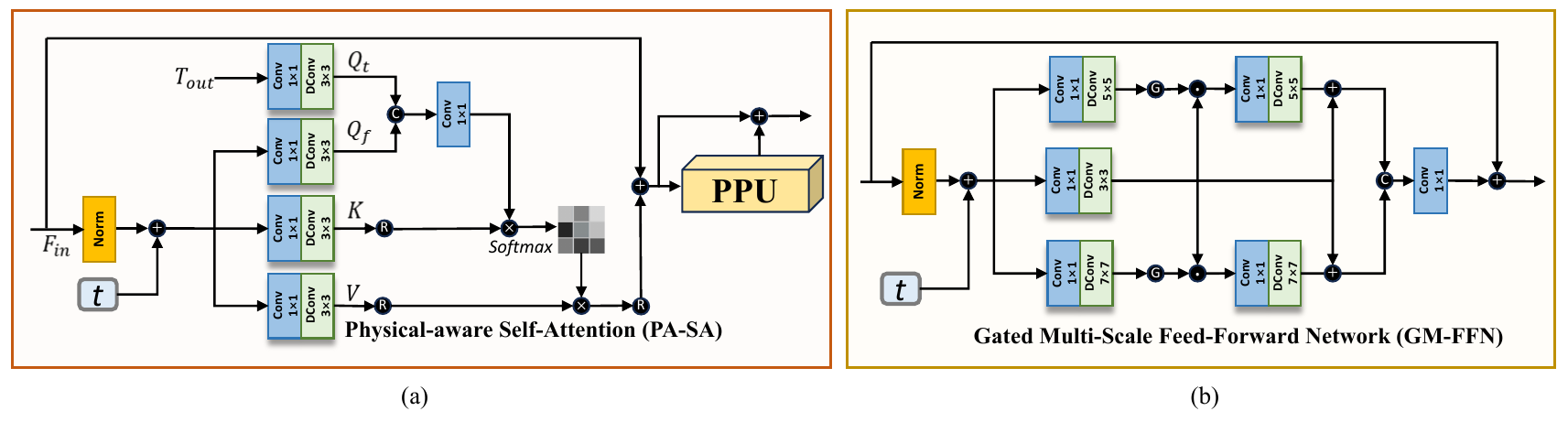}
	\vspace{-9pt}
	\vspace{-15pt}

	\caption{The detailed architecture of our designed (a) physics-aware self-attention  (PA-SA)  and (b) gated multi-scale feed-forward network  (GM-FFN).}
	
	\label{fig:block}
\end{figure*}
\begin{figure}[t]
	\centering
	\includegraphics[width=0.95\linewidth]{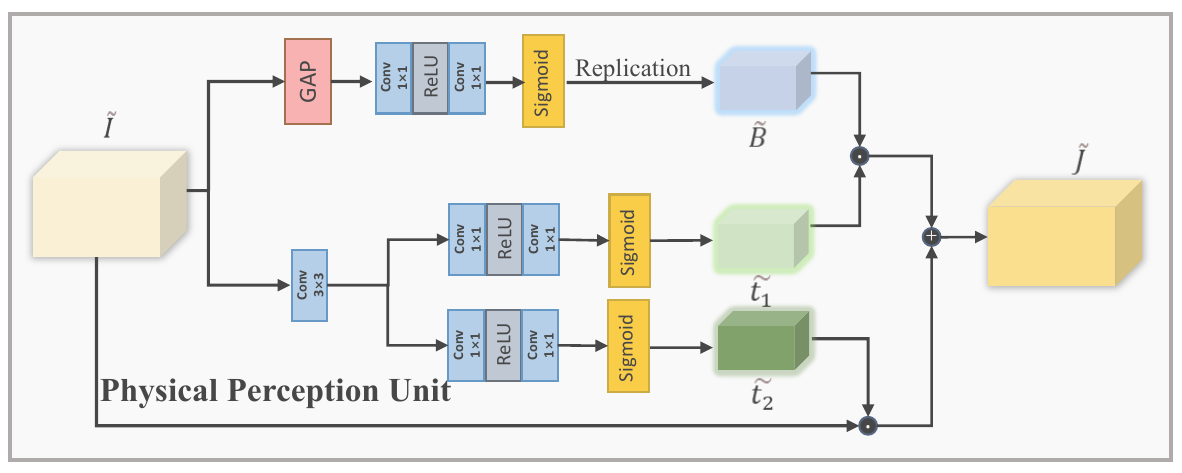}
	
	\caption{The detailed structure of the proposed physics perception unit (PPU).  To the best of our knowledge, \rev{it is the first method which is based on the interpretable physics model in the feature space for  challenging underwater image enhancement tasks.} 
	}
	\label{fig:ppu}
\end{figure}
\subsection{Implicit Neural Reconstruction}
Inspired by \cite{quan2023single,NeRD-Rain, yang2023implicit} based on implicit neural representations that can encode an image as a continuous function and learn a more robust feature representation, \rev{we further introduce Implicit Neural Representations (INRs) to learn robust degradation representations from diverse underwater images, aiming to reduce the difficulty of restoration for diffusion models. INRs train a multi-layer perceptron (MLP) by learning the following mapping :}
\begin{equation}F_\theta:\mathbb{R}^2\to\mathbb{R}^3,\end{equation}
\rev{$\mathbb{R}^2$ denotes the continuous 2D spatial coordinate domain $(x, y)$ and $\mathbb{R}^3$ denotes the corresponding RGB color space, i.e., each spatial location is mapped to its RGB color value. Specifically, following the setting of \cite{NeRD-Rain, yang2023implicit},  an underwater image  $I$ is transformed
to a feature map $\mathbf{E}\in\mathbb{R}^{H\times W\times C}$ with an image resolution of H × W.} Meanwhile, the position of each pixel is recorded in a relative coordinate set  $\mathbf{X}\in\mathbb{R}^{H\times W\times2}$, where
the value '2' means horizontal and vertical coordinates. As
shown in Figure \ref{fig:frame}, we fuse $X$ and $E$, and use $F_\theta$ to output image $\boldsymbol{I}_{{\mathrm{inr}}}$, which is expressed as:
\begin{equation}\boldsymbol{I}_{{\mathrm{inr}}}[x,y]=\boldsymbol{F}_\theta(\mathbf{E}[x,y],\mathbf{X}[x,y]),\end{equation}
where $[x, y]$ is the location of a pixel and $\boldsymbol{I}_{{\mathrm{inr}}}[x,y]$ is the generated RGB value. Due to the elimination of high-frequency components during rendering \cite{yang2023implicit},  we adopt periodic spatial encoding \cite{lee2022local} to project the pixel coordinates X into a higher dimensional space $\mathbb{R}^{2L}$ for better recovering high-frequency details. The encoding procedure is formulated as:
\begin{equation}\mathbf{X}^{\prime}=\gamma(\mathbf{X}),\end{equation}
\begin{equation}\gamma(x)=(\cdot\cdot\cdot,\sin(2^i x),\cos(2^i x),\cdot\cdot\cdot),\end{equation}
where $\gamma()$  represents a spatial encoding function. $i$ values from $0$ to $L-1$, and $L$ is a hyperparameter that determines the dimension value.  Given a GT of underwater image $GT$, The reconstruction loss of INR branch can be expressed as:
\begin{equation}\mathcal{L}_{INR}=||\boldsymbol{I}_{\mathrm{inr}}-GT||_{1}.\end{equation}

Finally, we fuse $I$ and $\boldsymbol{I}_{\mathrm{inr}}$ to obtain the condition $x_c$ of diffusion model, which can be expressed as:
\begin{equation}x_c=\boldsymbol{I}_{\mathrm{inr}}+ I .\end{equation}

\subsection{Physics-aware Diffusion Transformer}
\begin{figure}[t]
	\includegraphics[width=\linewidth]{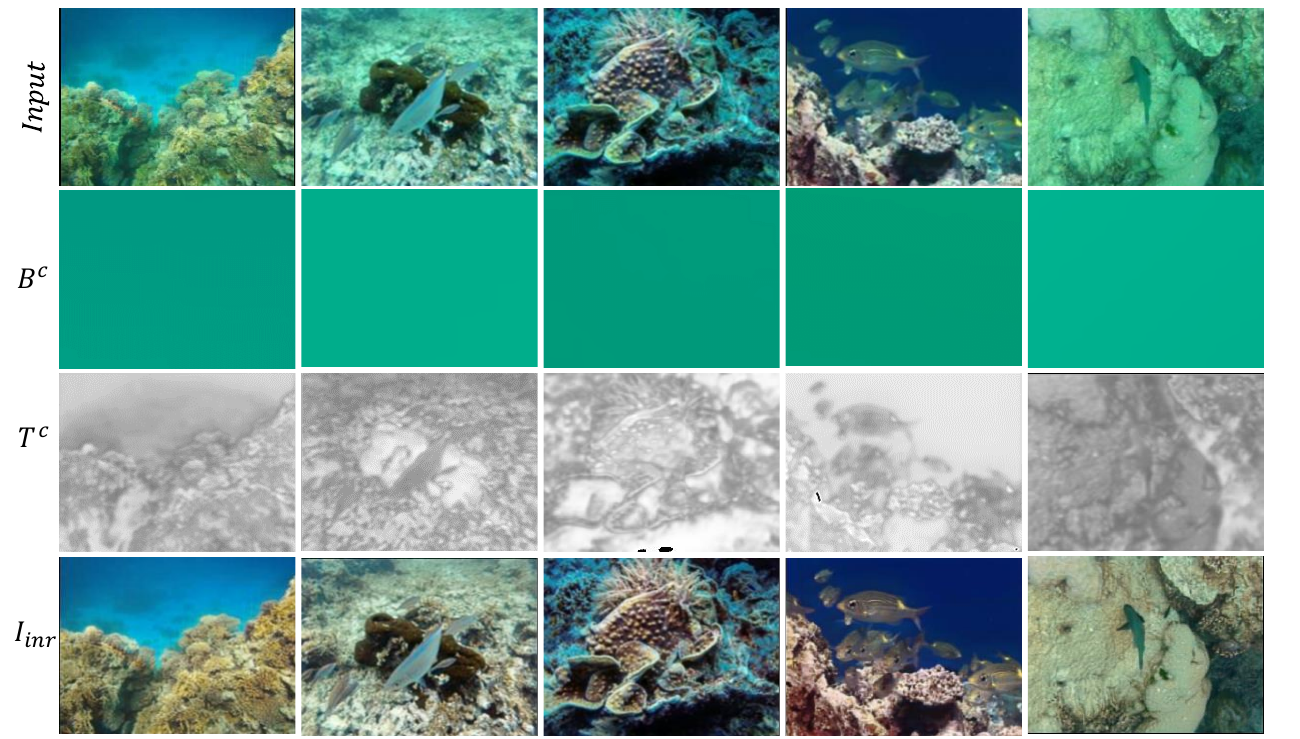}

	\caption{The visual results of the background scattered light $B^c$, the medium transmission map $T^c(x)$, and the implicit neural output $I_{inr}$. 
	}
 \label{fig:prior_visual}
\end{figure}

Physics-aware diffusion transformer (PDT) branch aims to exploit the extracted physics prior information from PPG branch for better guiding underwater image  restoration. Figure~\ref{fig:prior_visual} showcases the prior results provided by our PPG branch and INR branch. Figure~\ref{fig:frame} shows the overall of PDT branch, which is a U-shaped structure with \rev{physics-aware diffusion transformer (PDT) blocks and cross attention modules (CAM).} 
\rev{Given the output of the INR branch as the condition image} $x_{c} \in {R}^{ H \times W \times 3}$ and the background light $B^c \in {R}^{ H \times W \times 3}$  as a physical prior condition, we first obtain the noise sample $x_{t} \in {R}^{ H \times W \times 3}$ at time step $t$ according to the forward process, and concatenate $x_{t}$, $B^c$ and $x_{c}$ at channel dimension to obtain the input of PDT branch. Then, we obtain the embedding features $\mathbf{F} \in {R}^{  H \times W \times C}$ through convolution, where C means the
number of channel. $\mathbf{F}$ is encoded and decoded by PDT block, which consists of physics-aware self-attention  (PA-SA), physics perception unit and  gated multi-scale  feed-forward network  (GM-FFN). Note that, we introduce a physics perception unit (PPU) to further exploit the beneficial feature-level physics information.

\noindent \textbf{Cross Attention Module}. CAM aims to obtain a physics-aware interaction embedding, using which guides the diffusion process. We denote $T^c$ and $\mathbf{F}$ as the input of CAM. We use different linear projections to construct $Q$ and $K$ in CAM: $Q=Conv_{1\times1}(T^c),K=Conv_{1\times1}(\mathbf{F}).$
Similarly, $V$ of the transmission map $T^c$  can be obtained: $V=Conv_{1\times1}(T^c).$ The output feature map of CAM $T_{out}$  can then be obtained from the formula:
\begin{equation}T_{out}=Softmax(\frac{QK^T}{\sqrt{d_k}})V.\end{equation}

\noindent \textbf{Physics-aware Self-attention}. As is shown in Figure~\ref{fig:block} (a), we first embed the time embedding $\mathbf{t}$ into the input features $\mathbf{F}$ in our PA-SA: $\tilde{\mathbf{F}}=Norm(\mathbf{F})+\mathbf{t}$. We obtain query ($Q_f=\mathbf{W}_d\mathbf{W}_p\tilde{\mathbf{F}}$), key ($K=\mathbf{W}_d\mathbf{W}_p\tilde{\mathbf{F}}$), and value ($V=\mathbf{W}_d\mathbf{W}_p\tilde{\mathbf{F}}$) from the embedding input features $\mathbf{F}$. Meanwhile, we generate query ($Q_t=\mathbf{W}_d\mathbf{W}_p\mathbf{T_{out}}$) from the output feature map of CAM, where $\mathbf{W}_p$ and $\mathbf{W}_d$ respectively denote 1×1 point-wise convolution and 3×3 depth-wise convolution. We employ physics-aware query-driven attention strategy to obtain the final query $Q$ by fusing $Q_t$ and $Q_f$, which can be expressed as:
\begin{equation}
	Q=Conv_{1\times1}(concat(Q_f,Q_t)).
\end{equation}

We employ self-attention and get the output of PA-SA $\mathbf{\hat{F}}$:

\begin{equation}
	\hat{\mathbf{F}}=\mathbf{F}+softmax\left(\mathbf{Q}\mathbf{K}^T/\alpha\right)\cdot\mathbf{V},
\end{equation}		
where $\alpha$ is a learnable parameter. Consequently, PA-SA introduces physical guidance to fully exploit physics knowledge at the feature level and use self-attention mechanism to implicitly model the features of transmission map, which can help the diffusion model restore missing details and correct color distortion.

\noindent \textbf{Physics Perception Unit}. Exploring feature-level physics information \cite{zheng2023curricular,dong2020physics} achieved success in image dehazing tasks. Our proposed physics perception unit (PPU) aims to further exploit the beneficial feature-level physics information for underwater image enhancement, and learn richer physical features. Our proposed PPU is based on the simplified atmospheric model Eq. (5),  as
shown in Figure~\ref{fig:ppu}. We can reformulate the physics model to represent the clear image J, which can be obtained by:
\begin{equation}
	J^c(\mathbf{x})=I^c(\mathbf{x})\frac{1}{T^c(\mathbf{x})}+B^c(1-\frac{1}{T^c(\mathbf{x})}).
\end{equation}

Let $k$ denote a feature extractor(the filter kernel in a deep network). Then extracting features via the feature extractor k, Eq. (17) can be reformulated as follows:
\begin{equation}
	k\otimes J=k\otimes(I\odot\frac{1}{t})+k\otimes B(1-\frac{1}{t}),
\end{equation}
where $\otimes$ denotes the convolution operator and $\odot$ denotes the element-wise product operation. Using the matrix-vector form for algebraic operations, Eq. (26) can be express as:

\begin{equation}
	\boldsymbol{KJ}=\boldsymbol{K}\boldsymbol{T}_{1}\boldsymbol{I}+\boldsymbol{K}\boldsymbol{T}_{2}\boldsymbol{B},
\end{equation}
where $\boldsymbol{K}, \boldsymbol{J}, \boldsymbol{I}, \boldsymbol{B}, \boldsymbol{T}_{1}$ and $ \boldsymbol{T}_{2}$ respectively denote the matrix-vector forms of $k, J, I, B, \frac{1}{t}$ and $ (1-\frac{1}{t})$. Note that the diagonal vector of the diagonal matrix $\boldsymbol{T}_{1}$ corresponds to the vectorized form of $\frac{1}{t}$. 

Then, we can decompose the matrix $\boldsymbol{K}\boldsymbol{T}_{1}$ and $\boldsymbol{K}\boldsymbol{T}_{2}$ into product
of two matrices $\boldsymbol{F}_{1}\boldsymbol{K}$ and $\boldsymbol{F}_{2}\boldsymbol{K}$. Eq. (19) can be express as:
\begin{equation}
	\begin{aligned}\boldsymbol{KJ}
		&=\boldsymbol{F}_{1}(\boldsymbol{K}\boldsymbol{I})+\boldsymbol{F}_{2}(\boldsymbol{K}\boldsymbol{B}).\end{aligned}
\end{equation}
The Eq. (20) presents the relation between the clear image and the underwater image in a deep feature space. Due to the strong representation ability of deep neural networks, we can design a feature-level physics perception unit (PPU) and adopt deep networks to approximate the features corresponding to $\boldsymbol{F}_{1}$, $\boldsymbol{F}_{2}$ and $\boldsymbol{K}\boldsymbol{B}$. We denote $\tilde{\boldsymbol{B}}$ as an approximation of the features $\boldsymbol{K}\boldsymbol{B}$ that correspond to the background light,  and $\tilde{\boldsymbol{t}_{1}}$ and $\tilde{\boldsymbol{t}_{2}}$  as  approximations of $\boldsymbol{F}_{1}$ and $\boldsymbol{F}_{2}$, which are associated with the transmission map. Furthermore, $\boldsymbol{K}\boldsymbol{I}$ and $\boldsymbol{KJ}$ can be viewed as a latent underwater image $\tilde{\boldsymbol{I}}$  and its corresponding latent clear image $\tilde{\boldsymbol{J}}$, respectively. According to Eq. (20), the $\tilde{\boldsymbol{J}}$ can be express as:
\begin{equation}
	\tilde{\boldsymbol{J}}=\tilde{\boldsymbol{t}_{1}}\odot\tilde{\boldsymbol{I}}+\tilde{\boldsymbol{t}_{2}}\odot\tilde{\boldsymbol{B}}.
\end{equation}

Since the background light is usually assumed to be homogeneous, we can use global average pooling (GAP) to obtain  the main information for $\tilde{\boldsymbol{B}}$ in the feature space. $\tilde{\boldsymbol{B}}$ can be express:
\begin{equation}
	\tilde{\boldsymbol{B}}=H(\sigma(\mathrm{Conv}(\mathrm{ReLU}(\mathrm{Conv}(\mathrm{GAP}(\tilde{\boldsymbol{I}}))))))))),
\end{equation}
where $\sigma$ is the Sigmoid function, H denotes a replication operation.

As the transmission map is non-homogeneous, we cannot apply GAP for the approximations of $\tilde{\boldsymbol{t}_{1}}$ and $\tilde{\boldsymbol{t}_{2}}$. $\tilde{\boldsymbol{t}_{1}}$ and $\tilde{\boldsymbol{t}_{2}}$ can be obtained by 
\begin{equation}
	\tilde{\boldsymbol{t}_{1}}, \tilde{\boldsymbol{t}_{2}} =\sigma(\mathrm{Conv}(\mathrm{ReLU}(\mathrm{Conv}(\mathrm{Conv}(\tilde{\boldsymbol{I}})))))),
\end{equation}

\noindent \textbf{Gated Multi-scale  Feed-forward Network}. \rev{Finally, we design a GM-FFN for local feature aggregation.}  As is shown in Figure~\ref{fig:block} (b),  we employ GD-FFN-like structure \cite{zamir2022restormer} to construct our feed-forward network. In order to expand the receptive field, we employ multi-scale kernel depth-wise convolutions. GM-FFN adopts GELU to ensure the flexibility of feature aggregation.
\new{
{\flushleft{\textbf{ Discussion on Prior Multi-scale Works}.}}
MB-TaylorFormerv2 \cite{jin2025mb} introduces multi-scale patch embedding in transformer for image restoration, incorporating varying receptive field sizes, flexible receptive field shapes, and multi-level semantic information.
ESTINet \cite{zhang2022enhanced} proposes a feature-level multi-scale fusion module between the encoder and decoder for video deraining.
DDMSNet \cite{zhang2021deep} presents a multi-stage coarse-to-fine multi-scale CNN framework for snow removal, achieving progressive restoration.
GridFormer \cite{wang2024gridformer} adopts a pixel-level multi-scale transformer architecture for image restoration, effectively leveraging complementary multi-scale information in the pixel space.
Other works \cite{zhang2018adversarial, zhang2021benchmarking,zhao2025zero,ren2024fast,ZHAO2026112300} also explore multi-scale information fusion and utilization at both the pixel and feature levels. In contrast, our work investigates multi-scale information utilization within the feed-forward network (FFN) of transformers, enhancing the FFN’s ability to capture multi-scale features.
}

\begin{figure*}[t]
	\centering
	\includegraphics[width=1\linewidth]{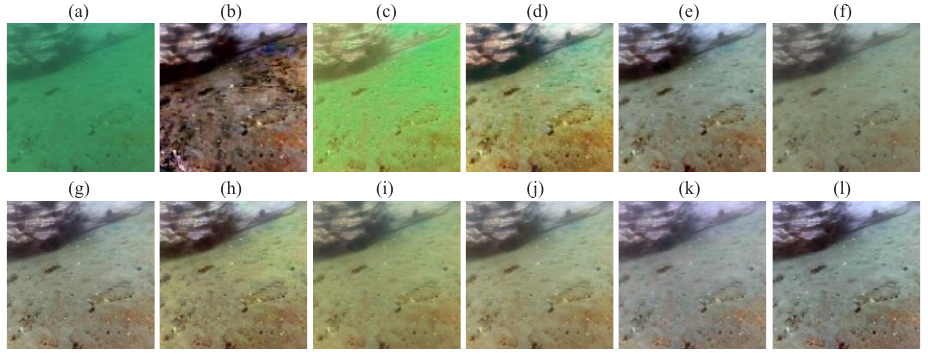}
	\caption{Qualitative comparison with other SOTA methods on the LSUI dataset. (a) input, (b) UIEWD,(c) UWCNN, (d) WaterNet, (e) UIEC2Net, (f)
Ucolor, (g) GUPDM, (h) CECF, (i) DM-water, (j) HCLR, (k) SS-UIE, (l) PA-Diff.} 
	
	\label{fig:lsui}
\end{figure*}
\begin{figure*}[t]
	\centering
	\includegraphics[width=1\linewidth]{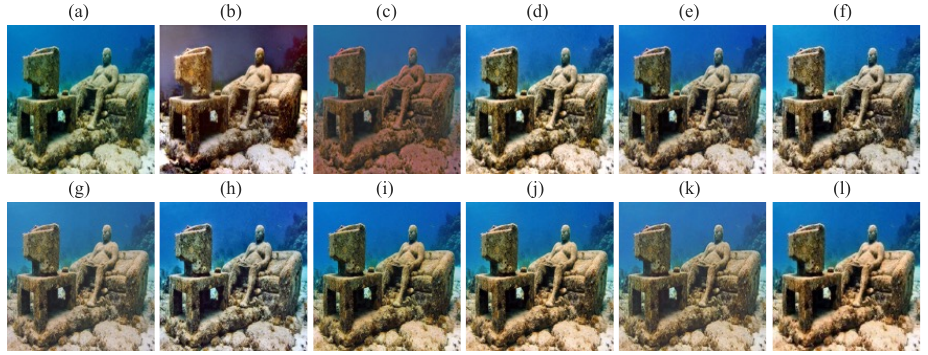}
	\caption{Qualitative comparison with other SOTA methods on the UIEB dataset. (a) input, (b) UIEWD,(c) UWCNN, (d) WaterNet, (e) UIEC2Net, (f)
Ucolor, (g) GUPDM, (h) CECF, (i) DM-water, (j) HCLR, (k) SS-UIE, (l) PA-Diff.} 
	
	\label{fig:uieb}
\end{figure*}

\begin{table}
\centering
\caption{Implementation details of our PA-Diff.}

\begin{tabular}{ll}
\toprule
\textbf{Hyperparameter} & \textbf{Value} \\
\midrule
Optimizer & Adam \\
Learning Rate & $1 \times 10^{-4}$ (fixed) \\
Batch Size & 6 \\
Total Iterations & 1,050,000 \\
Input Resolution & $256 \times 256$ \\
Diffusion Timesteps & 2,000 \\
Beta Schedule & Linear ($1 \times 10^{-6}$ to $1 \times 10^{-2}$) \\
UNet Inner Channel & 48 \\
Channel Multiplier & $[1, 2, 4, 8, 8]$ \\
Attention Resolution & 16 \\
Dropout & 0.2 \\
Norm Groups & 24 \\
GPU Hardware & NVIDIA RTX 3090 (24GB) \\
Training Time per Epoch & $\sim$8.6 minutes \\
Training Time per Iteration & $\sim$0.8 seconds \\
\bottomrule
\end{tabular}
\label{tab:implementation_details}
\end{table}

\begin{table*}[ht]
\centering
\caption{\rev{Quantitative comparison on the UIEB and LSUI datasets}. The best results are highlighted in bold and the second best results are underlined.}
\renewcommand{\arraystretch}{1.2}
\setlength{\tabcolsep}{4pt}
\begin{tabular}{l|cccccc|cccccc}
\toprule
\multirow{2}{*}{Method} & \multicolumn{6}{c|}{UIEB} & \multicolumn{6}{c}{LSUI} \\
& PSNR $\uparrow$ & SSIM $\uparrow$ & LPIPS $\downarrow$ & FID $\downarrow$ & UCIQE $\uparrow$ & UIQM $\uparrow$ & PSNR $\uparrow$ & SSIM $\uparrow$ & LPIPS $\downarrow$ & FID $\downarrow$ & UCIQE $\uparrow$ & UIQM $\uparrow$ \\
\midrule
UIEWD \cite{MaO22}        & 13.68 & 0.696 & 0.437 & 174.129 & 0.607 & 4.132 & 14.45 & 0.658 & 0.429 & 101.326 & 0.547 & 4.369 \\
UWCNN  \cite{abs-1807-03528}      & 13.94 & 0.716 & 0.335 & 91.048  & 0.551 & 4.082 & 16.59 & 0.706 & 0.361 & 61.202  & 0.551 & 4.103 \\
WaterNet  \cite{PengZB23}   & 22.76     & 0.906    & 0.129    & 42.432      & 0.615    & 4.341    & 20.78     & 0.804    & 0.234   & 37.900      & 0.569    & 4.349    \\
UIEC2Net \cite{WangGGY21}    & 20.04 & 0.818 & 0.178 & 49.071  & 0.606 & 4.369 & 24.25 & 0.854 & 0.142 & 37.425  & 0.591 & 4.307 \\
Ushape  \cite{LiGRCHKT20}      & 18.15 & 0.833 & 0.210 & 68.359  & 0.555 & 4.235 & 23.66 & 0.845 & 0.158 & 29.051  & 0.573 & 4.351 \\
Ucolor \cite{LiAHCGR21}      & 20.08 & 0.830 & 0.184 & 73.245  & 0.587 & 4.236 & 23.98 & 0.864 & 0.166 & 35.279  & 0.566 & 4.235 \\
GUPDM  \cite{mu2023generalized}      & 19.90 & 0.877 & 0.157 & 41.825  & 0.573 & 4.226 & 24.48 & 0.869 & 0.119 & 29.360  & 0.578 & 4.380 \\
CECF \cite{CongGH24}        & 22.20 & 0.841 & 0.135 & 32.370  & \underline{0.619} & \underline{4.374} & 26.58 & \underline{0.886} & 0.115 & \textbf{19.636}  & 0.590 & 4.355 \\
DM-water \cite{TangKI23} & 22.78 & 0.894 & 0.113 & 29.249  & 0.598 & 4.217 & \underline{26.95} & 0.881 & 0.143 & 21.433  & 0.589 & 4.257 \\
WF-Diff \cite{Zhaowfdiff} & 22.84 & 0.898 & 0.114 & 28.482  & 0.609 & 4.293 & 26.07 & 0.884 & 0.121 & 20.846  & 0.587 & 4.369\\
HCLR  \cite{ZhouSLJZLZF24}       & \underline{22.87} & \textbf{0.919} & \underline{0.112} & \underline{27.864}  & 0.607 & 4.292 & 26.33 & 0.880 & \underline{0.108} & 27.211  & \underline{0.593} & \textbf{4.408} \\
SS-UIE \cite{peng2025adaptive}       & 21.98 & \underline{0.907} & 0.121 & 36.479  & 0.589 & 4.263 & 24.41 & 0.840 & 0.164 & 26.196  & 0.575 & 4.318 \\
\midrule
PA-Diff      & \textbf{23.47} & 0.897 & \textbf{0.106} & \textbf{23.631}  & \textbf{0.621} & \textbf{4.385} & \textbf{27.25} & \textbf{0.889} & \textbf{0.103} & \underline{20.463}  & \textbf{0.594} & \underline{4.393} \\
\bottomrule
\end{tabular}

\label{tab:uieb_lsui_comparison}
\end{table*}

\begin{table*}[ht]
\centering
\caption{Quantitative comparison on EUVP, U45, and Challenge subsets. The best results are highlighted in bold and the second best results are underlined.}
\renewcommand{\arraystretch}{1.2}
\setlength{\tabcolsep}{4pt}
\begin{tabular}{l|cccccc|cc|cc}
\toprule
\multirow{2}{*}{Method} & \multicolumn{6}{c|}{EUVP} & \multicolumn{2}{c|}{U45} & \multicolumn{2}{c}{Challenge} \\
& PSNR $\uparrow$ & SSIM $\uparrow$ & LPIPS $\downarrow$ & FID $\downarrow$ & UCIQE $\uparrow$ & UIQM $\uparrow$ & UCIQE $\uparrow$ & UIQM $\uparrow$ & UCIQE $\uparrow$ & UIQM $\uparrow$ \\
\midrule
UIEWD \cite{MaO22}        & 13.93 & 0.623 & 0.452 & 91.712  & 0.504 & 4.063 & 0.510 & 4.057 & 0.550 & 3.641 \\
UWCNN \cite{abs-1807-03528}       & 17.50 & 0.692 & 0.341 & 55.767  & 0.570 & 4.157 & 0.534 & 4.075 & 0.535 & 3.671 \\
WaterNet \cite{PengZB23}  & 18.43     & 0.770    & 0.308    & 39.888      & 0.585    & 4.182    & 0.552    & 4.055    &  0.579    &  3.839    \\
UIEC2Net \cite{WangGGY21} & 24.04 & 0.818 & 0.180 & 29.755  & 0.594 & 4.248 & 0.575 & 4.105 & 0.581 & 3.935 \\
Ushape \cite{LiGRCHKT20}  & 25.53 & 0.818 & 0.139 & 24.914  & 0.587 & 4.238 & 0.567 & 4.139 & 0.530 & 3.820 \\
Ucolor \cite{LiAHCGR21}   & 25.68 & \underline{0.871} & 0.159 & 26.114  & 0.586 & 4.240 & 0.560 & 4.014 & 0.542 & 3.683 \\
GUPDM \cite{mu2023generalized}    & 25.18 & 0.848 & \underline{0.114} & \underline{20.744}  & 0.590 & \textbf{4.399} & 0.566 & \underline{4.144} & 0.543 & 3.827 \\
CECF \cite{CongGH24}      & \underline{28.02} & 0.861 & 0.126 & 22.630  & 0.596 & 4.252 & 0.585 & 4.128 & 0.575 & 3.899\\
DM-water \cite{TangKI23} & 26.45 & 0.852 & 0.183 & 24.484  & \underline{0.598} & 4.266 & \underline{0.592} & 4.139 & \underline{0.586} & 3.919 \\
WF-Diff \cite{TangKI23} & 27.24 & 0.851 & 0.128 & 21.517  & 0.592 & 4.269 & 0.584 & 4.137 & 0.584 & 3.926 \\
HCLR \cite{ZhouSLJZLZF24} & 27.08 & 0.850 & 0.117 & 25.114  & 0.591 & 4.288 & 0.584 & 4.117 & 0.562 & 3.792 \\
SS-UIE \cite{peng2025adaptive}  & 26.00 & 0.824 & 0.160 & 24.829  & 0.587 & \underline{4.353} & 0.584 & 4.105 & 0.557 & \textbf{3.970} \\
\midrule
PA-Diff                       & \textbf{29.12} & \textbf{0.880} & \textbf{0.101} & \textbf{18.769}  & \textbf{0.599} &  4.273 & \textbf{0.593} & \textbf{4.147} & \textbf{0.586} & \underline{3.954} \\
\bottomrule
\end{tabular}
\label{tab:euvp_u45_challenge}
\end{table*}

\begin{figure*}[t]
	\centering
	\includegraphics[width=1\linewidth]{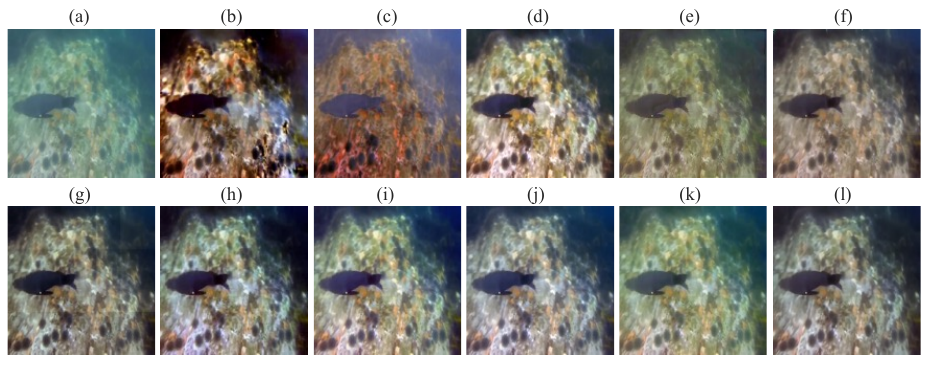}
	\caption{Qualitative comparison with other SOTA methods on the Challenge dataset. (a) input, (b) UIEWD,(c) UWCNN, (d) WaterNet, (e) UIEC2Net, (f)
Ucolor, (g) GUPDM, (h) CECF, (i) DM-water, (j) HCLR, (k) SS-UIE, (l) PA-Diff.} 
	\label{fig:realworld}
\end{figure*}

\begin{figure*}[t]
	\centering
	\includegraphics[width=1\linewidth]{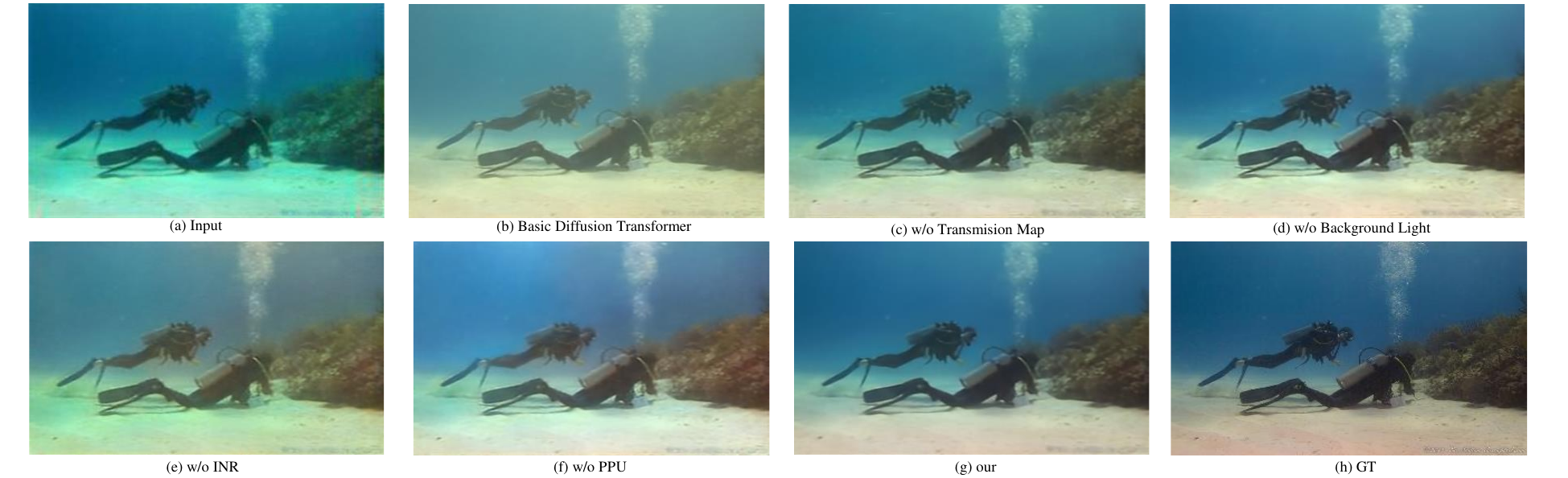} 
 \\ 
	\vspace{5pt} 
	\includegraphics[width=1\linewidth]{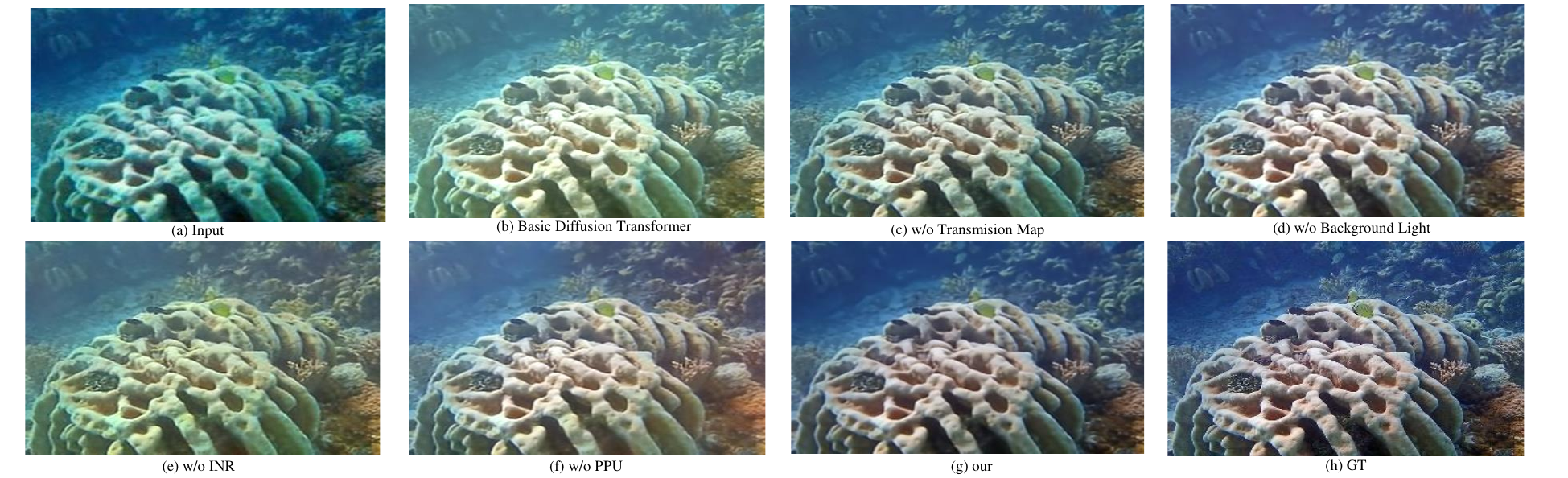} 
	\caption{Visual analysis with ablation study on the LSUI datasets.} 
	\label{fig:analysis}
 \vspace{-9pt}
\end{figure*}

\begin{figure}[t]
	\includegraphics[width=\linewidth]{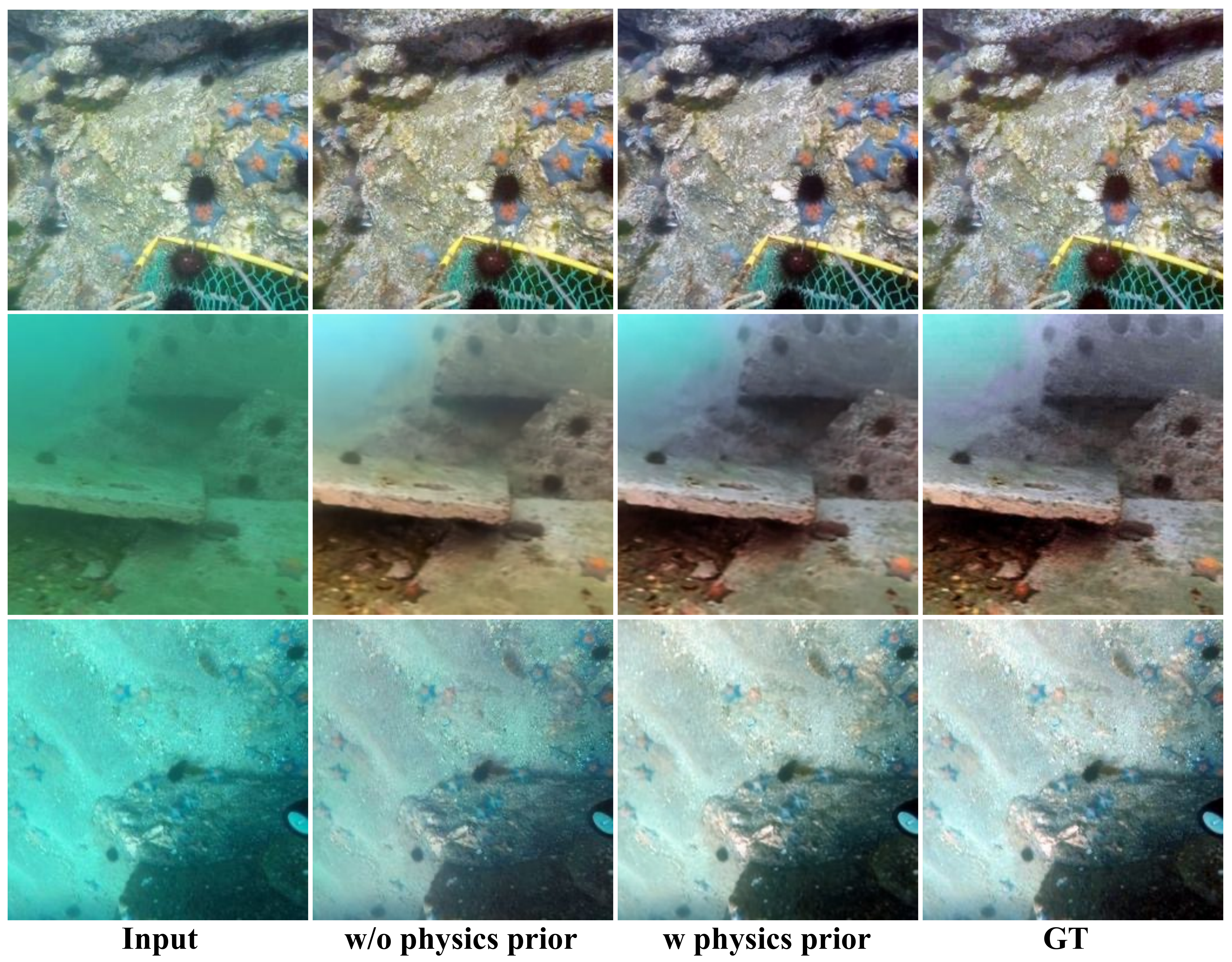}
	\caption{ \new{Comparative analysis of diffusion-based underwater image restoration with and
without physical priors.}}
 \label{fig:physics}
 \vspace{-9pt}
 \vspace{-9pt}
 \end{figure}
 
\begin{figure}[t]
	\includegraphics[width=\linewidth]{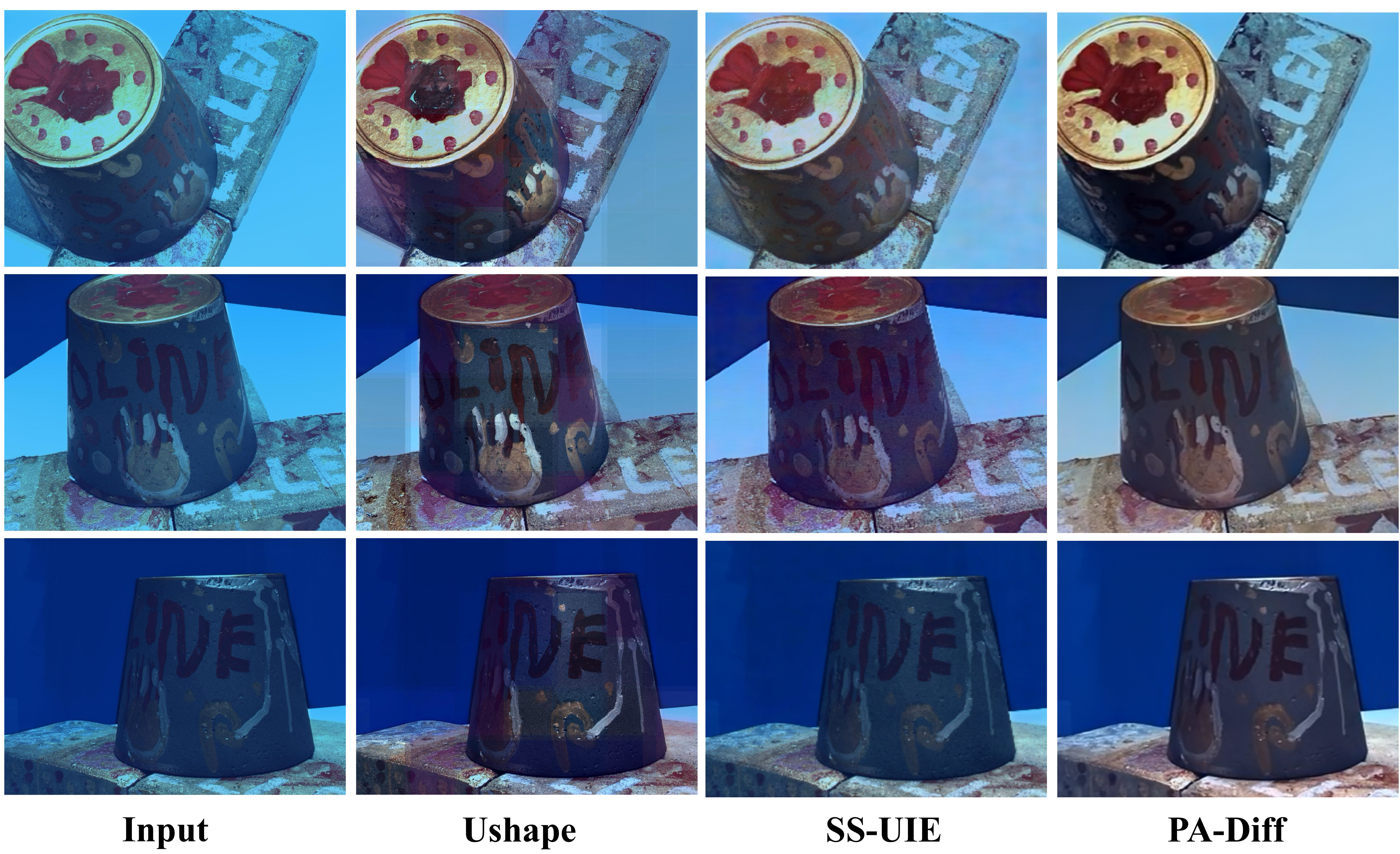}
	
	\caption{ \new{Multi-view visual comparison on UwMVS real-world test set \cite{11128539}.}
	}
 \label{fig:multiview}
	\vspace{-9pt}
\end{figure}

\begin{figure}[t]
	\includegraphics[width=\linewidth]{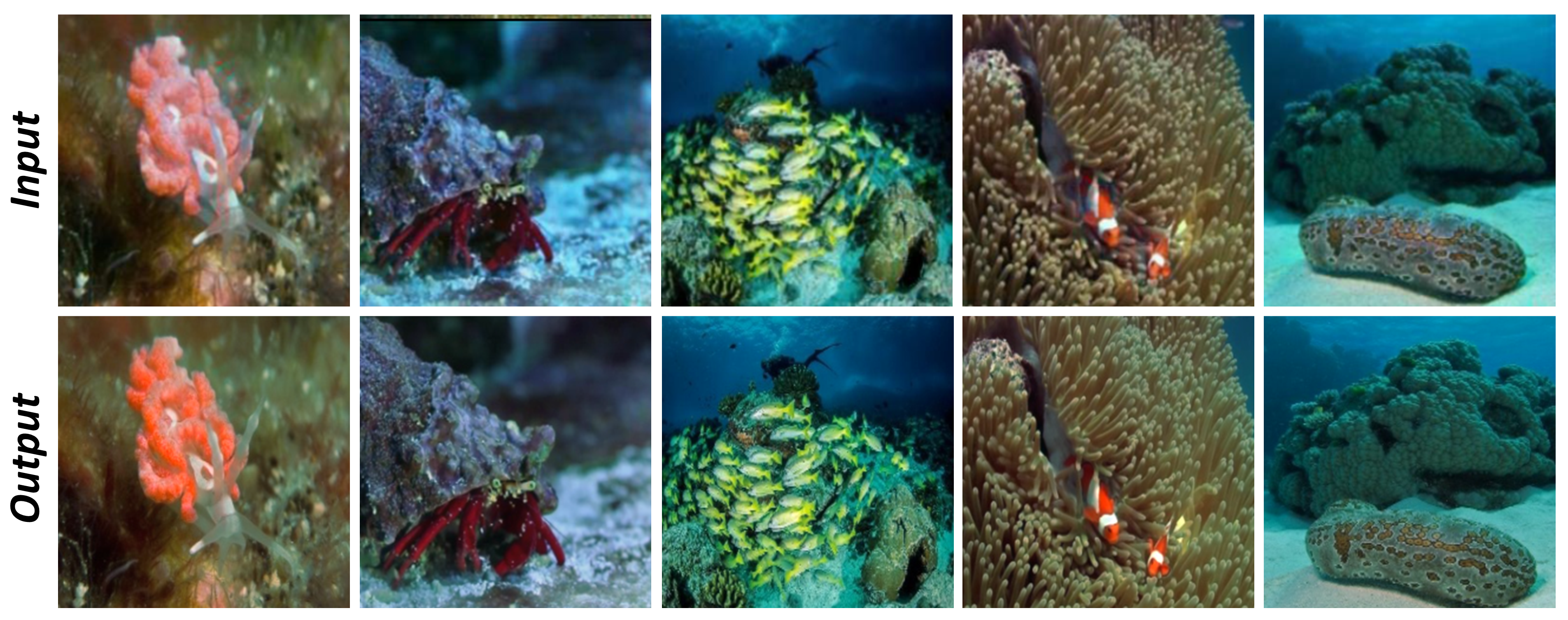}

	\caption{ Limitations. The output images refer to the reconstruction images of  PPG branch. 
	}
 \label{fig:Limitations}
	
\end{figure}

\begin{figure}[t]
	\includegraphics[width=\linewidth]{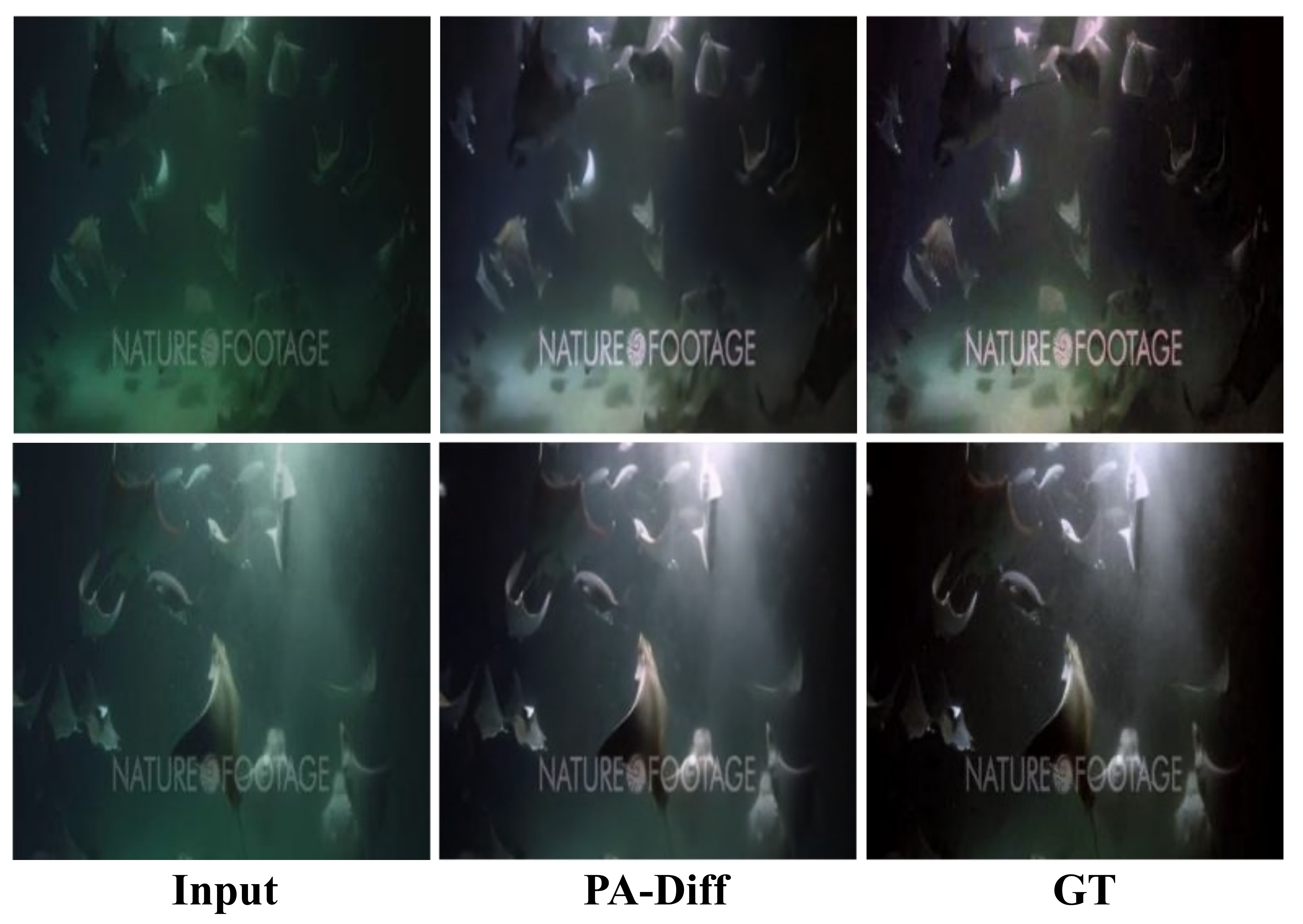}
	
	\caption{ \new{Visual results under extreme low-light conditions.}}
	\label{fig:lowlight}
 \vspace{-9pt}
\end{figure}
\section{Experiments}
\subsection{ Experimental Settings}
\noindent \textbf{Implementation Details}. Our network, implemented using PyTorch 1.7, underwent training and testing on an NVIDIA GeForce RTX 3090 GPU. We employed the Adam optimizer with $\beta_{1}= 0.9$ and $\beta_{2}= 0.999$. The learning rate was established at 0.0001. During the training stage, to balance the batch size and image size, \rev{the batch size and image resolution are set to 6 and  256 × 256, respectively.} The number of training iterations reached one million.  The pixel values of the image are normalized to [-1,1]. The time step of the diffusion model is set to 2000.  
\rev{A skip sampling strategy is employed with 10 sampling steps to balance enhancement performance and inference runtime.} \new{ Table \ref{tab:implementation_details} shows the key details regarding the implementation of our PA-Diff.}

\noindent \textbf{Datasets}. We utilize the  UIEB dataset \cite{LiGRCHKT20}  and the LSUI dataset \cite{PengZB23} for training and evaluating our model. \rev{The UIEB dataset comprises 890 underwater images with corresponding labels.}  Out of these, 800 images are allocated for training, and the remaining 90 are designated for testing.  LSUI dataset contains  4279 underwater images and their corresponding high-quality images. Compared with UIEB, LSUI contains diverse underwater scenes, object categories and deep-sea and cave images. \rev{In the paper, the LSUI dataset is randomly partitioned into 3879 images for training and 400 images for testing.} In addition, to verify the generalization, we use the reference based benchmark EUVP \cite{islam2020fast}, which contains 515 paired test samples, along
with the non-reference based benchmarks U45 \cite{li2019fusion} and Challenge \cite{LiGRCHKT20}, which contain 45
and 60 underwater images, respectively. 

\noindent \textbf{Comparison Methods}. We conduct a comparative analysis with seven SOTA UIE methods, namely UIECˆ2-Net \cite{WangGGY21}, Water-Net \cite{PengZB23}, UWCNN \cite{abs-1807-03528}, UIEWD \cite{MaO22}, U-color \cite{LiAHCGR21}, U-shape \cite{LiGRCHKT20}, GUPDM \cite{mu2023generalized}, CECF \cite{CongGH24}, DM-water \cite{TangKI23}, WF-Diff \cite{Zhaowfdiff}, HCLR \cite{ZhouSLJZLZF24}, and SS-UIE \cite{peng2025adaptive}. To ensure a fair and rigorous comparison, we utilize the provided source codes from the respective authors and adhere strictly to the identical experimental settings across all evaluations.

\noindent \textbf{Evaluation Metrics}. We employ  PSNR and SSIM \cite{WangBSS04} for quantitative comparisons at both pixel and structural levels. Additionally, LPIPS \cite{ZhangIESW18} and FID \cite{HeuselRUNH17} are utilized to evaluate perceptual performance.  \rev{For non-reference benchmarks U45 and Challenge, we introduce UIQM \cite{panetta2015human} and UCIQE \cite{yang2015underwater} to evaluate our method}.

\subsection{ Results and Comparisons}

\rev{Table \ref{tab:uieb_lsui_comparison} shows the quantitative results compared with different SOTA methods
on the UIEB and LSUI  datasets.} We mainly use PSNR, SSIM, LPIPS, FID, UCIQE and UIQM as our quantitative indices for UIEBD and LSUI datasets. 
The results show that our algorithm outperforms state-of-the art methods obviously, especially in terms of all metrics. \rev{Figure \ref{fig:lsui} and \ref{fig:uieb} present visual comparison results with other methods on the  LSUI and UIEB  datasets}, with selected examples randomly chosen from each dataset. Furthermore, Table \ref{tab:euvp_u45_challenge} presents a comparison using  pre-trained models on LSUI across multiple datasets, demonstrating the superior generalization capability of our method. Furthermore, in order to more effectively validate the generalization performance of our method on real-world images, we present enhancement results from the Challenge dataset in Figure \ref{fig:realworld}.
\rev{Our method consistently generates more natural and visually pleasing results}, strongly proving that PA-Diff has good generalization performance for real-world underwater images.

\begin{table}[t]
	
	\caption{ Ablation with application of diffusion technology.  }
	\centering
	\resizebox{1\linewidth}{!}{
		\begin{tabular}{l| c c c| c c  c}
			\specialrule{1.2pt}{0.2pt}{1pt}
			\multicolumn{1}{c}{Methods} & \multicolumn{3}{c}{LSUI~} & \multicolumn{3}{c}{EUVP}\\
			\midrule
			\multicolumn{1}{c}{Name}   & PSNR $\uparrow$ & SSIM $\uparrow$ & LPIPS $\downarrow$& PSNR $\uparrow$ & SSIM $\uparrow$ & LPIPS $\downarrow$\\
			\midrule

			w/o Diffusion &27.13 &0.883 & 0.127 & 27.75 & 0.868 & 0.153\\
			\midrule
            
			PA-Diff (Ours)   & \textbf{27.25}& \textbf{0.889} & \textbf{0.103} & \textbf{29.12}& \textbf{0.880} & \textbf{0.101}\\
			\specialrule{1.2pt}{0.2pt}{1pt}
	\end{tabular}}
	\label{tab:diff}
\end{table}

\begin{table*}
	\caption{\rev{Ablation study on the LSUI dataset. $T_c$ and $ B_c $ refer to transmission map and background light prior information in the PPG branch, respectively.} GBB means Gaussian blur module. INR represents our INR branch. PSE refers to the periodic spatial encoding \cite{lee2022local}, aiming to better recover high-frequency details. \rev{CAM and PPU denote the cross attention module and physics perception unit, respectively.}  $GM-FFN$ means multi-scale operation in the feed-forward network.}
	\centering
	\label{tab:abla}
	  \resizebox{\textwidth}{!}{
	\begin{tabular}{c|ccc|cc|cc|cc|cc}
		\specialrule{1.2pt}{0.2pt}{1pt}
		
		Method  &$ B_c $ &$T_c$ &GBB & INR &PSE & CAM & PPU &$GM-FFN$&$GD-FFN$  & PSNR$\uparrow$ & SSIM$\uparrow$        \\\midrule
		
		A&$\times$           & $\times$             & $\times$ &$\times$         &$\times$ &$\times$           & $\times$             & $\times$         &$\checkmark$  & 25.14 & 0.852       \\
		\midrule
		B&$\checkmark$ &  $\times$                     & $\times$         &$\times$ & $\times$   &$\times$           & $\times$             & $\times$         &$\checkmark$&25.67 & 0.857    \\

		C&$\checkmark$         &  $\checkmark$                &  $\times$              &	$\times$   &	$\times$         & $\times$             & $\times$         &$\times$ &$\checkmark$ & 25.73 & 0.865      \\
		D&$\checkmark$        &  $\checkmark$              & $\checkmark$     & $\times$      & 	$\times$           & $\times$             & $\times$         &$\times$ &$\checkmark$&25.94   & 0.868    \\
		
		\midrule
		E &$\checkmark$   & $\checkmark$        &  $\checkmark$             &  $\checkmark$     &  	$\times$           & $\times$             & $\times$         &$\times$ &$\checkmark$ & 26.47    &   0.873          \\
		
		F &$\checkmark$  & $\checkmark$        &  $\checkmark$             &  $\checkmark$     & 	$\checkmark$            & $\times$             & $\times$         &$\times$ &$\checkmark$  &26.51   &   0.872          \\
		\midrule
		G &$\checkmark$   & $\checkmark$        &  $\checkmark$             &  $\checkmark$     &  	$\checkmark$          & $\checkmark$             & $\times$         &$\times$ &$\checkmark$ & 26.79    &   0.880          \\
		
		H &$\checkmark$  & $\checkmark$        &  $\checkmark$             &  $\checkmark$     & 	$\checkmark$            & $\checkmark$            & $\checkmark$      &$\times$ &$\checkmark$  &27.04    &   0.883         \\
		\midrule
		Ours &$\checkmark$  & $\checkmark$        &  $\checkmark$             &  $\checkmark$     & 	$\checkmark$            & $\checkmark$            & $\checkmark$      &$\checkmark$ &$\times$  &\textbf{27.25}    &   \textbf{0.889}          \\

		\specialrule{1.2pt}{0.2pt}{1pt} 
		
	\end{tabular} }

\end{table*}

\begin{table}
	\caption{Comparison results of inference speed and Params. }
	\label{tab:speed}
	\centering
	\scalebox{0.9}{
		\begin{tabular}{c|cccc|c}
			\toprule
			
			Method &WaterNet & U-color  & Ushape&DM-water &  PA-Diff        \\\midrule
			
			Inference time(s) &0.61 & 2.75  & 0.03  & 0.14  &   0.23   \\
			Params(M)  & 24.81  &157.4  & 65.60  & 10.69 & 41.76     \\
			
			
			\bottomrule
	\end{tabular}}
	
	
\end{table}
\begin{table}
	\caption{{Params} analysis of each branch in PA-Diff. }
	\vspace{-9pt}
	\label{tab:Params}
	\centering
	\scalebox{0.9}{
		\begin{tabular}{c|ccc|c}
			\toprule
			
			Method   & PPG branch &INR branch &PDT branch &  PA-Diff (all)        \\\midrule
			
			Params(M)  & 0.0302  & 0.1487  & 41.31   & 41.76     \\
			
			\bottomrule
	\end{tabular}}

\end{table}
\vspace{-5pt}
\subsection{ Ablation Study }
{\flushleft{\textbf{ Effects of Diffusion Technology}}.} Firstly, to validate the value of the diffusion technique within our framework, Table \ref{tab:diff} presents the ablation study on the application of diffusion technology. The proposed model with diffusion technique achieves better results, demonstrating the significant performance improvements it brings to the model.

{\flushleft{\textbf{ Analysis of Model Complexity.}}} \rev{Table \ref{tab:speed} shows a comparison of inference speeds and model parameters with the previous methods}, indicating that our method demonstrates competitive model parameters. Table \ref{tab:Params} presents the network parameters for each branch, clearly indicating that the PPG and INR branches, as auxiliary networks for prior generation, have a minimal number of parameters and exhibit low complexity. \rev{However, we employ a transformer-based denoising network in the PDT branch, resulting in  relatively large parameters.} Therefore, our future endeavors will concentrate on investigating lighter-weight denoising networks.

{\flushleft{\textbf{Impact of Different Framework}.}}
To evaluate the impact of each strategy on the diffusion model, we conduct a comprehensive ablation study. Table \ref{tab:abla} shows the ablation results on the LSUI dataset, and Figure \ref{fig:prior_visual} presents the visual results of the prior knowledge.  Model A means that we just employ the most basic diffusion transformer structure. Firstly, we investigate the influence of the PPG branch on the diffusion process. Compared with model A, model B, C and D achieve better results, \rev{suggesting that physical prior knowledge helps the diffusion model better restore underwater images.} Particularly, Model B exhibits a remarkable advancement in terms of PSNR metric, suggesting that the transmission map $T_c$ plays a highly effective role in guiding the diffusion process. 
Subsequently, we explore the impact of the INR branch on the performance of the diffusion model. Model E outperforms D, suggesting that INR branch can improve the performance and reduce the difficulty of restoration for diffusion models. Model F outperforms E, implying that the periodic spatial encoding can enhance the representation ability of INR branch. 
Finally, Ours achieves the best performance, \rev{proving that all our proposed components are effective for UIE tasks.}

{\flushleft{\textbf{Analysis and Discussion}.}}
We conduct an exhaustive analysis of the individual contributions of each core component to the holistic performance by visually presenting the results of ablation studies, as illustrated in Figure \ref{fig:analysis}. Here, Model b represents the employment of the fundamental diffusion transformer architecture alone, serving as our baseline. \rev{Compared to the full model}, the results of Model b show that the visual effects obtained by using only the basic diffusion model can be unrealistic. Models c and d are designed to investigate the influence of incorporating prior physical knowledge—embodied by the transmission map  $T_c$ and background light information $B_c$, respectively—into the full model framework. Compared to model d, the results for model c are worse, mainly in terms of poorer recovery in some areas and less good colour recovery, suggesting that the transmission map $T_c$ we introduced plays a very important role.  Model e, on the other hand, focuses on elucidating the impact of integrating an Implicit Neural Representation (INR) branch.  \rev{Compared to the full model}, the results of Model e show that the visual effects obtained by using only the basic diffusion model can be bad, suggesting that INR branch can improve the performance and reduce the difficulty of restoration for diffusion models.  Lastly, Model F is pivotal in examining the role of the PPU module within the full model context. The visual evidence gathered unmistakably highlights the crucial role each segment of our proposed model plays in achieving optimal performance. 

\new{
{\flushleft{\textbf{Impact of Physical Priors}.}}
As illustrated in Figure \ref{fig:physics}, we show a comparative analysis of diffusion-based underwater image restoration with and without physical priors. The results demonstrate that, in the absence of physical priors, the diffusion model exhibits noticeable deviations in both global color correction and local detail reconstruction, particularly under severely degraded underwater conditions. In contrast, incorporating physical priors provides a robust guidance signal throughout the diffusion process, effectively  alleviating error accumulation during reverse denoising. This design enables the diffusion model to better align with the underlying degradation mechanism, thereby improving its generalization capability and restoration robustness.
}

{\flushleft{\textbf{The Plug-and-play Capability of PPU}.}}
Furthermore, \rev{
we  explore the application of the PPU in other  models.} Table \ref{tab:ppu} showcases the adaptability of PPU across different methods, demonstrating its universal  potential across diverse methods to exploit the beneficial feature-level physics information.

\new{
{\flushleft{\textbf{Impact of Different PPU Structure}.}}
We conduct an in-depth analysis of the key design choices in the proposed PPU. Specifically, we examine two critical physical parameters, $\tilde{\boldsymbol{t}_{1}}$ and $\tilde{\boldsymbol{t}_{2}}$, which are closely related to the transmission map. As shown in Table \ref{tab:PPUstruc}, removing either parameter leads to a significant performance degradation, validating the effectiveness of the current PPU design. Furthermore, we include comparisons with alternative feature-level physics modeling modules, such as the PDU used in C2PNet \cite{zheng2023curricular}. The results demonstrate that our PPU consistently achieves superior performance, indicating that the proposed PPU is more suitable for underwater image enhancement.}

\new{
{\flushleft{\textbf{Choice of Attention in the PPG Branch}.}}
We perform ablation studies on the specific attention mechanisms adopted in the PPG branch, as reported in Table  \ref{tab:attention}. Variants that remove either channel attention or spatial attention are evaluated. The results show a consistent, albeit slight, performance drop when either component is removed, supporting the effectiveness of our attention design.}

\new{
{\flushleft{\textbf{ A Lightweight Variant of Our
PA-Diff}.}}
we design a lightweight variant of our model by reducing the channel width and the number of modules in the transformer-based PDT. As reported in Table \ref{tab:light}, the proposed lightweight PA-Diff (Mini) achieves performance comparable to the full model across all evaluation metrics, while using only about one quarter of the parameters of the full model.
Moreover, when compared with DM-Water under a similar parameter budget, our lightweight model consistently delivers superior results, further validating the effectiveness and scalability of the proposed framework.}

\new{
{\flushleft{\textbf{Extention on Downstream Tasks}.}} 
As shown in Figure~\ref{fig:multiview}, we provide a multi-view visual comparison using the UwMVS real-world test set \cite{11128539}, which further validate the effectiveness of our proposed method in improving reconstruction quality.}

\begin{table}
	\caption{Ablation study with the PPU on UIEBD. }
	\label{tab:ppu}
	\centering
	\scalebox{0.92}{
		\begin{tabular}{cccc}
			\toprule
			
			Method &Unet & DW-water &Ushape        \\\midrule
			
			w/o PPU(PSNR/SSIM)  &17.41/0.7769  &22.78/0.894  &18.15/0.833      \\
			w PPU(PSNR/SSIM)  &19.64/0.8157  &22.93/0.897  & 20.54/0.848    \\
			
			\bottomrule
	\end{tabular}}
\end{table}

\begin{table}
	
	\caption{ \new{Ablation with the choice of PPU structure on the LSUI dataset.} }
	
	\centering
	\resizebox{1 \linewidth}{!}{
		\begin{tabular}{l c c c c c  c}
			\specialrule{1.2pt}{0.2pt}{1pt}
			\multicolumn{1}{c}{ } & \multicolumn{6}{c}{LSUI~}\\
			\midrule
			\multicolumn{1}{c}{Methods}   & PSNR $\uparrow$ & SSIM $\uparrow$ & LPIPS $\downarrow$ & FID $\downarrow$ & UCIQE $\uparrow$ & UIQM $\uparrow$\\
			\midrule
			w/o $\tilde{\boldsymbol{t}_{1}}$  &  26.31 &  0.880  &  0.130 & 24.61   &  0.590 &   4.306 \\
   w/o $\tilde{\boldsymbol{t}_{2}}$  & 25.88  &  0.876  & 0.135   & 33.96   & 0.585  &  4.232  \\
   PDU \cite{zheng2023curricular}  &  26.94 &  0.878  & 0.126  &  22.76  &   0.593 & 4.318  \\
			\midrule
			PA-Diff   & \textbf{27.25}& \textbf{0.889} & \textbf{0.103} & \textbf{20.46}& \textbf{0.594} & \textbf{4.393}\\
			\specialrule{1.2pt}{0.2pt}{1pt}
	\end{tabular}}
	\label{tab:PPUstruc}
\end{table}

\begin{table}
	
	\caption{ \new{Ablation with the attention mechanisms of the PPG Branch on the LSUI dataset. CA and SA denote channel attention and spatial attention, respectively.} }
	
	\centering
	\resizebox{1 \linewidth}{!}{
		\begin{tabular}{l c c c c c  c}
			\specialrule{1.2pt}{0.2pt}{1pt}
			\multicolumn{1}{c}{ } & \multicolumn{6}{c}{LSUI~}\\
			\midrule
			\multicolumn{1}{c}{Methods}   & PSNR $\uparrow$ & SSIM $\uparrow$ & LPIPS $\downarrow$ & FID $\downarrow$ & UCIQE $\uparrow$ & UIQM $\uparrow$\\
			\midrule
			w/o CA  &  27.11 &  0.885  &  0.110 & 22.14   &  0.592 &   4.372 \\
   w/o  SA  &  27.08 &  0.879  & 0.104  &  21.83  &   0.589 & 4.381  \\
			\midrule
			PA-Diff   & \textbf{27.25}& \textbf{0.889} & \textbf{0.103} & \textbf{20.46}& \textbf{0.594} & \textbf{4.393}\\
			\specialrule{1.2pt}{0.2pt}{1pt}
	\end{tabular}}
	\label{tab:attention}
\end{table}

\begin{table}
	\caption{ \new{Comparison  with lightweight variants on the LSUI dataset.}}
	
	\centering
	\resizebox{1\linewidth}{!}{
		\begin{tabular}{l c c c  c  c}
			\specialrule{1.2pt}{0.2pt}{1pt}
			\multicolumn{1}{c}{ } & \multicolumn{5}{c}{LSUI~}\\
			\midrule
			\multicolumn{1}{c}{Methods}   & PSNR $\uparrow$ & SSIM $\uparrow$ & LPIPS $\downarrow$  & UIQM $\uparrow$ & Params(M) $\downarrow$ \\
			\midrule
			Ushape  &  23.66 &0.845& 0.158 &  4.351 & 65.60 \\
   DM-water  &  26.95& 0.881& 0.143   &    4.257& \textbf{10.69}  \\
			\midrule
   PA-Diff (Mini)   & \underline{27.18} & \underline{0.889} & \underline{0.123} &   \underline{4.378}& \underline{11.58} \\
			PA-Diff (Ours)   & \textbf{27.25}& \textbf{0.889} & \textbf{0.103} &  \textbf{4.393} & 41.76\\
			\specialrule{1.2pt}{0.2pt}{1pt}
	\end{tabular}}
	\label{tab:light}
\end{table}


{\flushleft{\textbf{Limitations}.}}
As illustrated in Figure~\ref{fig:Limitations}, there remains a discernible discrepancy between the reconstructed images from the PPG branch and the corresponding input images, indicating an existing gap between the physical information generated by our PPG branch and the ideal or ground-truth physical representation. \rev{Consequently, how to generate more accurate physical prior knowledge remains a challenge for future research.} \new{Regarding the extreme low-light conditions, we add a new set of qualitative results on the LSUI dataset. As shown in Figure~\ref{fig:lowlight}, while our method is able to achieve noticeable overall enhancement in terms of brightness and color consistency, the recovery of fine-grained details remains challenging due to severe information loss. Moreover, the performance under such extreme conditions is further constrained by the limited quality of the available data, as the ground-truth images themselves are often noisy or poorly illuminated. Thus, we also plan to explore self-supervised learning paradigms and the construction of higher-quality datasets to better address extreme low-light underwater scenarios.}

\section{Conclusion}
In this paper, we develop a novel UIE framework, namely PA-Diff. With utilizing physics prior information to guide the diffusion process, PA-Diff can obtain underwater-aware ability and model the complex distribution in real-world underwater scenes. To the best of our knowledge, \rev{it is the first  diffusion model based on physical perception for UIE tasks.} Our designed PPU is a plug-and-play universal module to exploit the beneficial feature-level physics information, which can be integrated into various deep learning network. \rev{PA-Diff shows SOTA performance on UIE task, and extensive ablation experiments validate the effectiveness of each proposed component.}

\bibliographystyle{IEEEtran}
\bibliography{refer}

\end{document}